\pdfoutput=1

\documentclass[10pt,twocolumn,letterpaper]{article}

\usepackage{wacv}
\usepackage{times}
\usepackage{epsfig}
\usepackage{graphicx}
\usepackage{amsmath}
\usepackage{amssymb}
\usepackage{booktabs}
\usepackage{breakcites}
\usepackage{booktabs}
\usepackage{adjustbox}
\usepackage{multirow}
\usepackage{makecell}
\usepackage{bbding}

\usepackage{multirow}
\usepackage{xspace}
\usepackage{enumitem}

\usepackage{pifont}% http://ctan.org/pkg/pifont
\newcommand{\cmark}{\ding{51}}%
\newcommand{\xmark}{\ding{55}}%

\usepackage{algorithmicx}
\usepackage[section]{algorithm}
\usepackage[numbered]{algo}
\usepackage{color}
\usepackage{array,xcolor,colortbl}
\usepackage{siunitx}
\usepackage{bbm}
\usepackage{authblk}
\usepackage[accsupp]{axessibility}

\def\wacvPaperID{929} % Enter the WACV Paper ID here
\wacvfinalcopy % *** Uncomment this line for the final submission

\definecolor{HL}{rgb}{0.9,0.9,0.9}

\ifwacvfinal
\usepackage[breaklinks=true,bookmarks=false]{hyperref}
\else
\usepackage[pagebackref=true,breaklinks=true,colorlinks,bookmarks=false]{hyperref}
\fi

\definecolor{Gray}{gray}{0.9}
\usepackage{footnote}
\usepackage[hang,flushmargin]{footmisc}
\makeatletter
\newcommand{\algorithmfootnote}[2][\footnotesize]{%
  \let\old@algocf@finish\@algocf@finish% Store algorithm finish macro
  \def\@algocf@finish{\old@algocf@finish% Update finish macro to insert "footnote"
    \leavevmode\rlap{\begin{minipage}{\linewidth}
    #1#2
    \end{minipage}}%
  }%
}

\begin{document}

%%%%%%%%% TITLE
\title{Towards Discriminative and Transferable One-Stage Few-Shot Object Detectors\vspace{-1em}}

% \author[1,2]{Karim Guirguis$^\dagger$\footnote{karim.guirguis@de.bosch.com}}
% \author[3]{Mohamed Abdelsamad \thanks{Equal Contribution}}
% \author[3]{George Eskandar}
% \author[3]{Ahmed Hendawy}
% \author[1]{Matthias Kayser}
% \author[3]{Bin Yang}
% \author[2,4]{Juergen Beyerer}

% \affil[1]{Robert Bosch GmbH, Germany}
% \affil[2]{Karlsruhe Institute of Technology, Germany}
% \affil[3]{University of Stuttgart, Germany}
% \affil[4]{Fraunhofer IOSB, Germany}

\author{
Karim Guirguis\thanks{Both authours have contributed equally to this work} $^{ 1,2}$\hspace{1mm}
Mohamed Abdelsamad$^{*3}$\hspace{1mm}
George Eskandar$^{3}$\hspace{1mm}
Ahmed Hendawy$^{3}$\hspace{1mm}
Matthias Kayser$^{1}$\\ \vspace{-5mm}
\and 
Bin Yang$^{3}$ \hspace{1mm}
Juergen Beyerer$^{2,4}$\\ 
{\normalsize Robert Bosch GmbH$^1$\thanks{karim.guirguis@de.bosch.com}}\hspace{1mm}
{\normalsize Karlsruhe Institute of Technology$^2$}
{\normalsize University of Stuttgart$^3$}
{\normalsize Fraunhofer IOSB$^4$}
\vspace{-2em}
}
% \author{Karim Guirguis\\
% Robert Bosch Corporate Research\\
% Renningen, Germany\\
% {\tt\small karim.guirguis@de.bosch.com}
% % For a paper whose authors are all at the same institution,
% % omit the following lines up until the closing ``}''.
% % Additional authors and addresses can be added with ``\and'',
% % just like the second author.
% % To save space, use either the email address or home page, not both
% \and
% Mohamed Abdelsamad\\
% University Of Stuttgart\\
% Stuttgart, Germany\\
% {\tt\small  mam1996@gmail.com}
% \and
% George Eskandar\\
% University Of Stuttgart\\
% Stuttgart, Germany\\
% {\tt\small  george.eskandar@iss.uni-stuttgart.de}
% \and
% Ahmed Hendawy\\
% University Of Stuttgart\\
% Stuttgart, Germany\\
% {\tt\small   ahmedmagdyahmed1996@outlook.com}
% \and
% Matthias Kayser\\
% Robert Bosch Corporate Research\\
% Renningen, Germany\\
% {\tt\small   matthias.ochs2@de.bosch.com}
% \and
% Bin Yang\\
% University Of Stuttgart\\
% Stuttgart, Germany\\
% {\tt\small   bin.yang@iss.uni-stuttgart.de}
% \and
% Juergen Beyerer\\
% Fraunhofer IOSB\\
% Karlsruhe, Germany\\
% {\tt\small   juergen.beyerer@iosb.fraunhofer.de}
% }
%\input{egrebuttal}

\clearpage 
\maketitle

\thispagestyle{empty}

%%%%%%%%% ABSTRACT
\begin{abstract}
\vspace{-1em}
Recent object detection models require large amounts of annotated data for training a new classes of objects. Few-shot object detection (FSOD) aims to address this problem by learning novel classes given only a few samples. While competitive results have been achieved using two-stage FSOD detectors, typically one-stage FSODs underperform compared to them. We make the observation that the large gap in performance between two-stage and one-stage FSODs are mainly due to their weak discriminability, which is explained by a small post-fusion receptive field and a small number of foreground samples in the loss function. To address these limitations, we propose the Few-shot RetinaNet (FSRN) that consists of: a multi-way support training strategy to augment the number of foreground samples for dense meta-detectors, an early multi-level feature fusion providing a wide receptive field that covers the whole anchor area and two augmentation techniques on query and source images to enhance transferability. Extensive experiments show that the proposed approach addresses the limitations and boosts both discriminability and transferability. FSRN is almost two times faster than two-stage FSODs while remaining competitive in accuracy, and it outperforms the state-of-the-art of one-stage meta-detectors and also some two-stage FSODs on the MS-COCO and PASCAL VOC benchmarks.
\end{abstract}
\vspace{-2em}
%%%%%%%%% BODY TEXT
\section{Introduction}

The scene understanding of the environment is crucial for autonomous systems, such as industrial robots or self-driving cars.
One of the main pillars of such a perception system is the object detection task. 
With the advent of deep learning and the availability of enormous annotated datasets, object detection methods are rapidly improving in terms of both efficiency and detection performance. 
However, current object detectors do not generalize well when learning novel unseen categories with limited data. 
To this end, few-shot learning (FSL) has been proposed to allow for improved knowledge transfer from base classes with abundant data to novel classes with a handful of examples. 
%FSL mimics the human cognitive ability to learn new concepts from a few examples. 
Originally, FSL was designed and applied to the image classification task~\cite{vinyals2016matching,snell2017prototypical,finn2017model,ravi2017optimization,sung2018learning,garcia2018few,qi2018low, antoniou2018data,hariharan2017low,wang2018low}.
Adapting FSL to object detection is considerably harder due to the challenging localization task and the more complex architecture of object detectors. 

Few-shot object detection (FSOD) approaches can be grouped into two main categories: transfer learning and meta-learning based methods.
Transfer learning based approaches~\cite{LSTD, TFA, MPSR, FSCE, defrcn} transfer knowledge from base classes by finetuning this pre-trained model on the novel categories without training it from scratch.
In contrast, meta-learning methods~\cite{DANA, CME, FsDetView, FSOD-RPN, MetaRCNN, FSRW, MetaDet, DCNET} strive to rapidly learn novel classes by leveraging class-agnostic representations extracted during training. 
%In each training episode in meta-FSOD, different tasks are being solved, where a task is defined as the localization of objects belonging to different classes in a query image, given a set of support images of the same classes. 
Most meta-learners can either learn to finetune or learn to compare. 
The former implies learning a better parameter initialization to adapt to new tasks in a few gradient steps. 
The latter aims to project features into an embedding space where the similarity between query and support can be effectively measured. 
By striving to minimize a meta-loss over various episodes, meta-learning can extract a prior knowledge that is transferable to a new task with limited data.

Although significant progress has been achieved in applying meta-learning to two-stage (sparse) object detectors, one-stage (dense) meta-detectors are understudied. 
There exists only a few one-stage few-shot object detectors~\cite{FSRW, once}, and even fewer dense meta-detectors~\cite{FSRW}, and their performance significantly lags. 
In this work, we are improving dense detectors in a few-shot setting because they are faster, lighter and more embedded-friendly. 
%Additionally, they were shown to outperform two-stage detectors, in an abundant data regime (e.g., EfficientDet~\cite{EfficientDet}). 
To this end, we first propose a simple method to evaluate and analyze dense object detectors: (1) how discriminative the detector is on the abundant base classes, which can be measured by the average precision (AP) on base classes (bAP), and (2) how transferable is the extracted knowledge to novel classes. To measure the latter, we propose to measure the ratio of AP on novel classes (nAP) to bAP: the higher the ratio, the more transferable the model is.
Using these evaluation metrics, we seek to find out the limiting factors in one-stage meta-detectors.
Our experiments show that the degraded performance on novel classes can be mainly attributed to the low discriminability. 
This is due to a direct application of meta-learning from image classification that ignores the nature of the object detection task. 

Thus, we develop a framework, named Few-Shot RetinaNet (FSRN) that addresses the common limitations.
Our contributions can be summarized as follows:
\begin{enumerate}[nolistsep]
    \item A multi-way support training strategy to increase the number of foreground detections per query image and provide a more robust learning signal to avoid vanishing gradients.
    \item An early multi-level fusion between support and query features prior to the class and box heads. This ensures that the post-fusion networks have a wide-enough local receptive field to cover the entire anchor area.
    \item Two augmentation techniques to enhance transferability: a multi-scale foreground data augmentation scheme and a new sampling of class prototype by using the normal distribution of the support shots features to simulate more shots.
\end{enumerate}
In our experiments, we show that our proposed FSRN outperforms state-of-the-art dense meta-detectors by a large margin and also many two-stage meta-detectors on the MS-COCO and PASCAL-VOC benchmarks, while being faster and lighter.
%Firstly,  Secondly, A multi-scale foreground data augmentation scheme enriches object scales while a higher weight in the focal loss is given to the foreground samples. Thirdly,  Finally, a max-margin loss aims to broaden the feature space of the detector and disentangle features of base and novel classes. To evaluate our model, we conduct extensive experiments to showcase the robust performance of the proposed model on the well-established MS-COCO and PASCAL-VOC benchmarks. FSRN outperforms state-of-the-art dense meta-detectors and achieves competitive results with many two-stage meta-detectors.

\section{Related Works}

\textbf{Object Detection.} Object detectors can be mainly categorized into two main groups: two-stage, or sparse, and one-stage, or dense, detectors. The former~\cite{R-CNN, FastR-CNN, FasterR-CNN} comprises a region of interest (RoI) generation step, where an object is most likely to be located. In Faster R-CNN~\cite{FasterR-CNN}, this is achieved by a region proposal network
(RPN), which classifies and refines the predicted proposals. Afterward, the proposals are pooled with the backbone feature map and fed into the classification and localization heads. On the other hand, one-stage detectors~\cite{YOLOv1,SSD, YOLOv2, RetinaNet, YOLOv3, YOLOv4, EfficientDet} omit the region proposals stage. YOLO~\cite{YOLOv1} is a single neural network that divides the input image into square grids, and for each grid, it directly regresses the bounding box coordinates and the class probability. Since then, multiple variants have followed~\cite{YOLOv2, YOLOv3, YOLOv4}, improving accuracy and speed. Most notably, RetinaNet~\cite{RetinaNet} was proposed to solve the foreground-background class imbalance problem encountered in dense detectors through the use of a focal loss and a Feature Pyramid Network (FPN) \cite{fpn} on top of a ResNet backbone. In this work, we transform the RetinaNet to a meta-detector for FSOD.  

\begin{table*}[t!]
\centering
\footnotesize
\setlength{\tabcolsep}{0.5em}
\caption[Proposed transferability metrics of some works]{{10-shot detection performance on MS-COCO~\cite{coco} dataset.} Viewing the Attention-RPN in the FSDO-RPN~\cite{FSOD-RPN} as a stand-alone dense meta-detector, we analyze the performance using the proposed evaluation protocol. The outcomes indicate that the Attention-RPN of the sparse FSOD-RPN meta-detector shows a poor performance similar to the dense Meta-YOLO~\cite{FSRW}. This throws a light on the significant gap between the RPN and the final detection head of FSOD-RPN. 
}
\label{tab:transferability}
\adjustbox{width=0.9\linewidth}{
\begin{tabular}{l|cccc|cccc|cccc}
\toprule
\multirow{2}{*}{Method} & \multicolumn{4}{c|}{Base Performance} & \multicolumn{4}{c|}{Novel Performance} & \multicolumn{4}{c}{Transferability} \\ 
&  bAP     & bAP50     & bAP75    &  bAR   & nAP   & nAP50     & nAP75    & nAR    & PT   & PT50  &  PT75     & RT   \\ \midrule
MetaYOLO~\cite{FSRW}  & 13.8 & - &  - & 15.5 & 5.6 & - & - & 14.4 &  0.40 & - & - & 0.93 \\
%ONCE~\cite{once} &  22.9 & - &  - & 29.9 & 5.1 & - & - & 9.5 &  0.22 & - & - & 0.32 \\ 
FSOD-RPN~(RPN-only)~\cite{FSOD-RPN} & 5.54 & 13.35 & 3.65 & 21.23 & 0.98 & 3.40 & 0.31 & 11.84 &  0.18 & 0.25 & 0.08 & 0.55 \\
FSOD-RPN~\cite{FSOD-RPN}  &  24.26 & 38.04 & 26.44 & 40.56 & 11.95 & 22.37 & 11.79 & 30.84 & 0.49 & 0.59 & 0.45 & 0.76 \\ 
\bottomrule
\end{tabular}}
\vspace{-2em}
\end{table*}
\textbf{Few-Shot Object Detection.} Recent FSOD approaches are either transfer learning or meta-learning based. Firstly, the transfer learning based few-shot detectors~\cite{LSTD, RepMet, TFA, MPSR, defrcn} strive to transfer knowledge from base classes to novel classes via finetuning on the latter. On the other hand, meta-detectors extract knowledge across different detection tasks to generalize better on new tasks. Meta-detectors can be grouped into two main approaches: learn to finetune~\cite{MetaDet} and learn to measure~\cite{RepMet, FSRW, MetaRCNN, FSOD-RPN, FsDetView, CME, DCNET, DANA, MetaDetR}. The former seek to learn category-agnostic parameters that enable the learning of novel category-specific weights on the new task~\cite{MetaDet}. In contrast, the latter models perform an exemplar search at the instance level given a support set with few images. This search is accomplished through a feature fusion between query and support features. However, the models vary in 3 aspects: \textit{where} to fuse, \textit{how} to fuse and the training strategy. MetaYOLO~\cite{FSRW} is a single-stage meta-detector that performs feature fusion directly before the detection head. Similarly, the fusion in two-stage detectors like MetaRCNN~\cite{MetaRCNN} and FsDetView~\cite{FsDetView} occur at the instance-level after the RPN. FSOD-RPN~\cite{FSOD-RPN} adds feature fusion before the RPN to filter the proposals by class category. Moreover, it is the only framework that employs a two-way contrastive training strategy. Overall, the feature fusion takes the form of an attention mechanism employing a Hadamard product while it consists of subtraction, multiplication, and concatenation operations in FsDetView\cite{FsDetView}. Among the meta-detectors above, only MetaYOLO~\cite{FSRW} is one-stage based, and it significantly lags the two-stage meta-detectors.

\section{Investigating Dense Meta-Detectors}
\label{sec:investigate}
\subsection{Problem Formulation}
In a FSOD setting, two datasets are utilized: a base dataset $\mathcal{D}_b$ containing abundant data of base classes $\mathcal{C}_b$, and a novel dataset $\mathcal{D}_n$ comprising a handful examples of novel classes $\mathcal{C}_n$.  Base and novel classes are disjoint (i.e., $\mathcal{C}_b \cap \mathcal{C}_n = \emptyset$). Formally, $\mathcal{D}_b=\{\ (x_i, y_i)\ |~y_i = \{(c_l,b_l)\}_{l=1}^{m_{i}}, c_l \in \mathcal{C}_{b}\}$, 
$\mathcal{D}_n=\{\ (x_i, y_i)\ |~y_i = \{(c_l,b_l)\}_{l=1}^{m_{i}}, c_{l} \in \mathcal{C}_{n}\}$, where $x \in \mathcal{X}$ is an input image, $y$ is the set of corresponding annotations per image, and $m_{i}$ is the number of annotations per image $x_i$. For each instance $l$, $c_l$ and $b_l$ are the class label and bounding box coordinates, respectively. The total number of images in $\mathcal{D}_n$ is much smaller than $\mathcal{D}_b$, $\vert\mathcal{D}_n\vert \ll \vert\mathcal{D}_b\vert$. In $\mathcal{D}_{n}$, for each class $c \in \mathcal{C}_n$, there are only $K$ annotations in total. 

Meta-learning based FSOD models are trained in an episodic manner. Each episode $e$ comprises an $N$-way-$K$-shot task $\mathcal{T} = \{\{\mathcal{S}^1, \dots, \mathcal{S}^N\}, \mathcal{Q}\}$. Each task is made up of a $N$ $K$-shot labeled support sets $\mathcal{S}^j$, where $\mathcal{S}^j = \{s_1, \dots, s_K \}$ is a support set for class $j$ and $\mathcal{Q}$ is a query image featuring instances of $N$ classes. Each support image $s$ is a close-up of an object of interest (i.e., cropped image via the bounding box). During training, for each task within an episode, the loss function considers the performance on the query image conditioned on the given support set. Not only does episodic training with non-overlapping tasks mimic the test-time, but also learns to classify and localize various objects in general rather than a specific set. 

\subsection{Analysis of Dense Meta-Detectors}
In order to find out where the performance gap between one-stage and two-stage meta-detectors happens, we propose to measure the performance by two simple metrics. First, we measure the AP on base classes which reflects the discriminability of the detector. Second, the ratios $PT = nAP / bAP$ and $RT=nAR / bAR$ are calculated, where AR is the average recall. PT denotes precision transferability, and RT denotes recall transferability. The second metric reflects the transferability of the model. An ideal meta-detector should have PT and RT equal to $1$, whereas low ratios indicate overfitting to the base classes. We start from the observation that the RPN in two-stage detectors is, in fact, a dense detector, and we report the aforementioned metrics on a two-stage meta-detector and its RPN as a stand-alone one-stage detector. Specifically, we choose FSOD-RPN~\cite{FSOD-RPN} because it employs a feature fusion before the RPN which generates class specific proposals and hence can be considered as a dense meta-detector. The results are reported in Table \ref{tab:transferability}, where we add MetaYOLO~\cite{FSRW} as a dense detector.

The results reveal that the RPN of the two-stage FSOD-RPN exhibits a poor performance similar to the one-stage MetaYOLO. There is a significant gap between the RPN and the last detection head of FSOD-RPN. Surprisingly, the difference manifests in the base classes, not just the novel classes, denoting low discriminability. We note that the transferability scores of MetaYOLO and the two-stage FSOD-RPN are close ($0.40$ vs $0.49$) as shown in Table \ref{tab:transferability}. In contrast, the transferability of the RPN of FSOD-RPN has dropped to the half, and we hypothesize that is mostly a byproduct of the low discriminability of the dense RPN.

We argue that there are multiple reasons for the low discriminability of dense detectors. First, the absence of a instance-level network in one-stage FSODs limits the receptive field and restricts the learning capacity post-fusion. Second, the learning signal in dense detectors is weaker than in sparse-detectors because most anchors are classified as background (negative examples) due to the query-set construction strategy that considers only a single class per image in each task. 

 \begin{figure*}[t!]
 \centering
 \includegraphics[width=.8\linewidth]{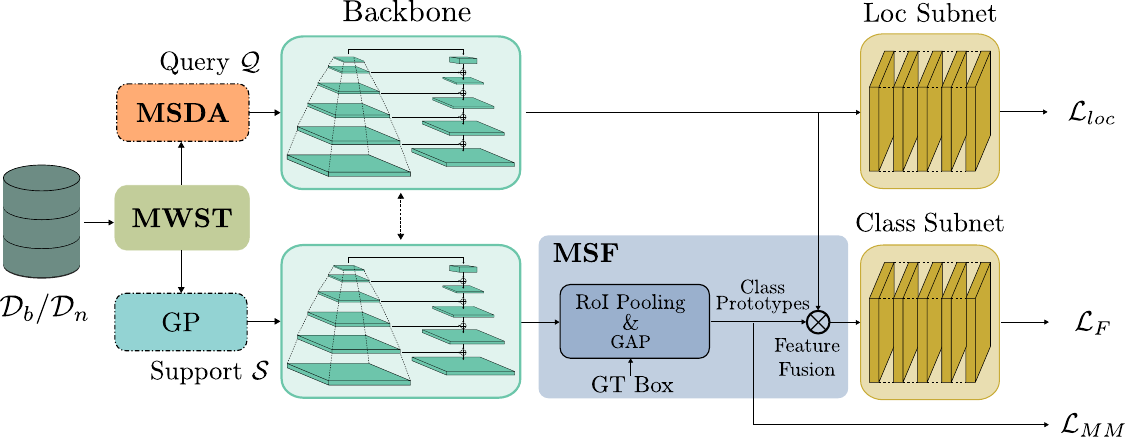}
 \caption{Overview of the our FSRN architecture. First, the multi-way support training strategy (MWST) constructs multi-way tasks per training episode, featuring multiple positive and negative classes. Thereby, more foreground anchors are sampled, resulting in improved discriminability. Next, a multi-scale feature fusion (MSF) module is used on top of the FPN that enables a wide receptive field covering the whole anchor area post-fusion. During meta-testing, an introduced multi-scale data augmentation scheme (MSDA) enriches the scale-space improving the discriminability for novel classes. Additionally, the proposed Gaussian Prototyping (GP) for improved class prototypes.} %
 \label{fig:architcture}%
 \vspace{-1.5em}
\end{figure*}
\section{Approach}

In this section, we present our approach Few-Shot RetinaNet (FSRN). It consists of 3 components: a multi-scale feature fusion (MSF), which allows for a wide receptive field covering the whole anchor area, a multi-way support training strategy (MWST) to increase the number of foreground samples enriching the learning signal, and a data augmentation strategy on both query and support images during meta-testing to enrich the data distribution.

\subsection{Architecture Overview}
As depicted in Fig. \ref{fig:architcture}, our proposed few-shot RetinaNet (FSRN) model extends the RetinaNet~\cite{RetinaNet} to a meta-detector. The architecture consists of 2 branches in a Siamese-like fashion, one for the query images and the other for the support images. Both branches share a common backbone which consists of a ResNet-50 and an FPN. In the support branch, the backbone is followed by a RoI-pooling operation to extract the relevant feature maps from the support images. Global average pooling (GAP) is then performed, followed by an averaging across the shots dimension to get the class prototypes. Next, the MSF module aggregates the query and class prototypes prior to the classification subnet, whereas the localization subnet only considers the query features.    

\subsection{Training Objective}

Similar to the original RetinaNet~\cite{RetinaNet}, we train using the focal loss:
\begin{equation}
    \mathcal{L}_F = - (\alpha~p_t~(1-p)^\gamma~\log(p) + (1-\alpha)~(1-p_t)~p^\gamma~\log(1-p)),
\end{equation}
where $p$ is the predicted class probability. $p_t$ denotes the ground truth label. $\alpha$ is a weighting factor, while $\gamma$ is a modulating parameter, which can be tuned to focus on the hard negative examples and reduce the loss contribution of easy ones. However, in the case of FSOD, we find that this training objective alone is not enough for the backbone to learn robust disentangled representations for the novel categories. To enhance discriminability and stabilize the training, we employ a max-margin loss inspired by~\cite{CME} aiming to minimize the intra-class variance while maximizing the inter-class variance. Formally, 
\begin{equation}
    \mathcal{L}_{MM} = \frac{ \sum_i^C \frac{1}{K} \sum_k^K ||v_{ik} - \mu_i||_2^2}{\sum_i^C \min_{j,j \neq i} ||\mu_i - \mu_j||_2^2},
\end{equation}
where $v_{ij}$ denotes the $k$-th prototype vector for class $i$, and $K$ is the number of prototype vectors. $\mu_i$ is the mean prototype for class $i$. $C$ is the total number of classes. To this end, the final training objective function can be denoted as follows:
\begin{equation}
    \mathcal{L} = \mathcal{L}_{F} + \mathcal{L}_{loc} + \lambda \mathcal{L}_{MM}, 
\end{equation}
where $\mathcal{L}_{loc}$ is the smooth $L1$-loss for the bounding box regression task. $\lambda$ is a scaling factor to balance the max-margin loss contribution compared to the classification and regression losses.

\begin{figure*}[t!]
 \centering
 \includegraphics[width=0.8\linewidth]{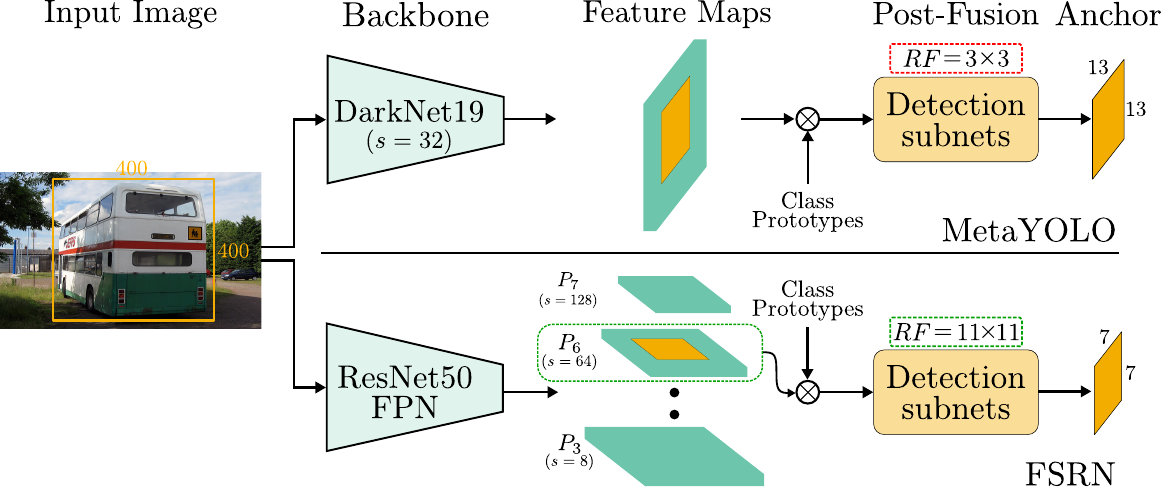}
 \caption{A depiction of the post-fusion network receptive field (RF) impact. To illustrate, we use an example query image from the MS-COCO~\cite{coco} dataset with an annotated bounding box of size $400\times400$. The upper part shows that a YOLOv2-based dense meta-detector~\cite{FSRW} suffers from a narrow receptive field that is is unable to cover the whole anchor area (i.e., $RF=3\times3 < 13\times13$). On the other hand, the proposed FSRN exploits the FPN~\cite{fpn} via the introduced MSF along with both deeper and wider post-fusion network to process the whole anchor area (i.e., $RF=11\times11 >7\times7$). }  %
\label{fig:receptive_field}%
\vspace{-1.5em}
\end{figure*}
\subsection{Early Multi-Scale Feature Fusion}
The experiments from Section~\ref{sec:investigate} have revealed that one of the reasons for limited discriminability correlates to the absence of a post-fusion network before the detection head. The fusion in meta-learning filters the global-level "class-agnostic" features learned by the backbone, resulting in "class-specific" features. Directly aggregating support and query features before the detector injects this "class-specific" information from the support branch, which the downstream layers cannot learn effectively because of their small receptive field and limited learning capacity. In two-stage FSOD-RPN, this loss of spatial information is alleviated by the presence of an RoI head that offers a wide-enough receptive field to learn instance-level features post-fusion. We hypothesize that the post-fusion receptive field should at least cover the area of the largest anchor size. In dense detectors, a naive solution is to increase the receptive field by cascading multiple layers between the fusion location and the detection head. However, a significant number of layers is necessary for the receptive field to cover the biggest anchor size, which renders the one-stage detector inefficient. 

The proposed MSF module is designed on top of the FPN to remedy this issue. The FPN by design limits the biggest anchor size to $10\times10$, which is easily covered by the downstream subnet, as shown in Figure~\ref{fig:receptive_field}. We fuse the features directly after the FPN. Specifically, support-level features are pooled from their corresponding level $p_l$ in the FPN, based on the groundtruth bounding box size. After spatial global averaging the extracted features from each support shot, the class prototype is computed by averaging across the $K$-support shots. Formally, the class prototype $\mu_c$ is computed as follows:
\begin{equation}
    \mu_c = \frac{1}{K}\sum_{k=1}^{K} \textit{GAP}(v_{ck}^{p_l}),
\end{equation}
where $v_{ck}^{p_l}$ is the the support feature of class $c$ from shot $k$ and the corresponding level $p_l$.
Finally, each class prototype attends the multi-level query features $f_Q^{p_l}$ through a Hadamard product operation to yield output features $f_o^{p_l}$ for each corresponding pyramid level $p_l$ as follows:
\begin{equation}
   f_o^{p_l} = \mu_c \odot f_{Q}^{p_l}.
\end{equation}
We opt for fusing features only prior to the classification subnets while directly feeding the extracted query features to the localization subnets without fusion to ensure that the localization task remains class-agnostic.

The choice of the fusion location allows for a deeper post-fusion network, which helps the backbone focus on global-level feature learning, while the subnets post-fusion learns the instance-level features. In order to foster the learning signal of the detection subnets, we increase the number of positive anchors per query image, by increasing the number of anchors per feature pixel from $9$ in the original RetinaNet to $15$. 

\subsection{Multi-Way Support Training Strategy}
\begin{figure*}[t!]
 \centering
 \includegraphics[width=.8\linewidth]{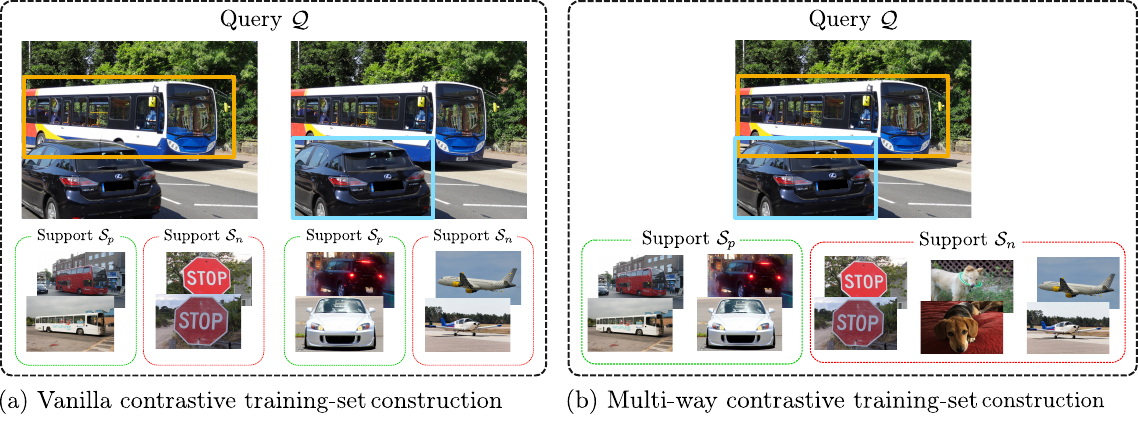}
 %\vspace{-1em}
 \caption{ The left image shows the query-support set construction in a contrastive-based settings for the FSOD-RPN~\cite{FSOD-RPN}. Here, one annotation per query image is sampled along with $K$ support shots from the same class annotation and $K$-shots from a random negative class. The MWST algorithm constructs a multi-way query-support set (right), where the query image can include multiple annotations.} %
 \label{fig:mwst}%
 \vspace{-1em}
\end{figure*}
%\vspace{-0.5em}
In meta-detection, the query-support set construction strategy is to usually sample all annotations in a query image belonging to single class $c$ along with $K$ support shots of the same class \cite{FSRW, FSOD-RPN, FsDetView, DANA}, as shown in Figure~\ref{fig:mwst}. This, in turn, limits each task per episode to a single-class detection. While the said strategy is suitable for image classification, object detection is a more challenging setting, where multiple class instances are present per query image. The binary query-support selection strategy leads to fewer foreground samples and, consequently, fewer positive anchors and fewer positive gradients available during the training. This aggravates an already existing problem in dense detectors, namely the overwhelming number of generated static anchors which contain background samples. Although the focal loss addresses the foreground-background imbalance problem, we observe that it does not entirely alleviate the issue for meta-detectors. To this end, we propose a multi-way support training strategy (MWST) as a remedy (Figure~\ref{fig:mwst}).         

Precisely, the query image is loaded with all its annotations for each task. A random class dropout is performed on the annotations, meaning when a class is dropped, all its corresponding annotations in the query image are removed. Next, $K$ support shots for each class are sampled. We limit the number of classes to $N$ per query image to limit the total number of support images needed and the associated computational cost. If the number of classes after dropout is smaller than $N$, negative classes are sampled in the support set $\mathcal{S}$. The proposed query-set construction algorithm Alg.~\ref{alg:support_algo} enables multi-class contrastive training, enriching the number of foreground objects to $\Bar{m}/2$, compared to $\Bar{m}/\Bar{c}$ in binary meta-detection, where $\Bar{m}$ is the average number of annotations per query image and $\Bar{c}$ denotes the average number of classes per query image. Moreover, the class dropout works as a data augmentation tool that simulates the random task sampling of a generic meta-learning paradigm and increases the cardinality of the query set from $\Bar{m} \times \vert \mathcal{D}_b \vert$ to $2^{\Bar{m}} \times \vert \mathcal{D}_b \vert$. The task where all classes are dropped is ignored.    

\setlength{\textfloatsep}{15pt}% Remove \textfloatsep
\begin{algorithm}[t!]
    \algorithmfootnote{$y_0$ denotes the initial value.}
	\begin{algo}{Query-specific support set $\mathcal{S}_i$ generator}
		{
			\small 
			\label{alg:support_algo}
			%x_i image vector 
			\qinput{Query element $\mathcal{Q}_i=(x_i,y_i)$, Support set $\mathcal{S}$, Set of classes $\mathcal{C}_i$ with instances in $x_i$, $N$ number of classes per task}
			\qoutput{Multi-way support set $\mathcal{S}_i$ for query $x_i$} 
		}   initialize $\mathcal{S}_i$ as empty list\\
		randomly drop classes from $y_i$\\ 
		\qfor every class $j$ in $y_i$ \\
		sample different $K$-shots from $\mathcal{S}^j$ \Comment{$S^j$ is the support set of class $j$.}\\ 
		add  to $\mathcal{S}_i$\qrof\\
		
		\qwhile $\mid \mathcal{S}_i \mid < N$ \\
		randomly select class $z$ from $\mathcal{C}_b \setminus \mathcal{C}_i$\label{algo_line:removal}\\
		sample different $K$-shots from $\mathcal{S}^z$\\
		add to $\mathcal{S}_i$\qelihw\\
		\qreturn $\mathcal{Q}_i, \mathcal{S}_i$
	\end{algo}
	\caption[Multi-way support set $s^i$]{Multi-way support set generation algorithm.}
    
\end{algorithm}
\vspace{-1em}
%$S^j$ is the support set of class $j$ in the whole training dataset.

\subsection{Multi-Scale Data Augmentation}

As shown in~\cite{MPSR}, during the meta-test phase, the limited novel data results in a sparse scale-space that can be divergent from the learned base distribution. Wu \etal~\cite{MPSR} have proposed a multi-scale positive samples refinement (MPSR) approach to address the scale variation problem by further exploiting the FPN. They added an auxiliary branch, which generates object pyramids of various scales and then refines the predictions accordingly. However, this incurs computational, time, and memory costs. Inspired by Wu \etal~\cite{MPSR}, we propose a multi-scale data augmentation module (MSDA) during meta-testing as an approximation of the multi-positive sample refinement approach. We approximate the refinement scheme by jittering the size of both the query and support images via a logarithmic-based scaling to cover all the FPN levels equally and weighting the foreground samples more via the $\alpha$ parameter in the focal loss. Empirically, $\alpha$ is increased to $\frac{\alpha + 1}{2}$, where $\alpha < 1$. For example, if $\alpha=0.5$ during meta-training, then we set $\alpha=0.75$ during the meta-testing phase. This reformulation provides a comparable performance without computational overhead during meta-testing.  

\begin{table*}[t!]
	\addtolength{\tabcolsep}{3pt}
	\begin{center}
	\caption{{ Evaluation of few-shot object detection on MS-COCO.} 
	We report the average precision and recall metrics for the 20 novel VOC classes with $K=5,10, 30$-shot. '-' denotes unreported results by the original paper. 
	}
	\label{tab:DetCoco}
	\scalebox{0.8}{
	\begin{tabular}{l | cccc | cccc | cccc}
	\toprule
	\multirow{2}{*}{Method} & \multicolumn{4}{c|}{5-Shot} & \multicolumn{4}{c|}{10-Shot} & \multicolumn{4}{c}{30-Shot} \\
	& AP & AP50 & AP75 & AR & AP & AP50 & AP75 & AR & AP & AP50 & AP75 & AR \\
	\midrule
	MetaYOLO~\cite{FSRW} &  - & - & - & - & 5.6 & 12.3 & 4.6 & 14.4 &  11.3 & 21.7 & 8.1 & 19.2\\
	%MetaDet-YOLO~\cite{MetaDet} &  - & - & - & - & 7.1 & 14.6 & 6.1 & 15.5 & 9.1 & 19.0 & 7.6 & 17.8\\
	ONCE~\cite{once} & - & - & - & - & 5.1 & - & - & 9.5 & - & - & - & - \\ 
	\rowcolor{HL}
	FSRN &  \textbf{8.7} & \textbf{16.1} &  \textbf{8.2} &  \textbf{27.5} & \textbf{15.8} & \textbf{26.4} & \textbf{15.9} & \textbf{36.0} & \textbf{18.1} & \textbf{30.5} & \textbf{39.4} & \textbf{39.6}\\
	\midrule
	%LSTD~\cite{LSTD} & - & - & - & - & 3.2 & 8.1 & 2.1 & 10.4 & 6.7 & 15.8 & 5.1 & 14.3  \\
	%MetaDet~\cite{MetaDet} & - & - & - & - & 7.1 & 14.6 & 6.1 & 15.5 & 11.3 & 21.7 & 8.1 & 19.2 \\
	Meta-RCNN~\cite{MetaRCNN} & - & - & - & - & 8.7 & 19.1 & 6.6 & 17.9 & 12.4 & 25.3 & 10.8 & 21.7 \\
 	TFA w/fc~\cite{TFA} & 8.4 & - & - & -  & 10.0 & 17.3 & 8.5 & - & 13.4 & 22.2 & 11.8 & - \\
 	TFA w/cos~\cite{TFA} & 8.3 & 13.3 & 6.5 & - & 10.0 & 17.1 & 8.8 & - & 13.7 & 22.0 & 12.0 & - \\
 	MPSR~\cite{MPSR}& - & - & - & - & 9.8 & 17.9 & 9.7 & 21.2 & 14.1 & 25.4 & 14.2 & 24.3 \\
 	FsDetView~\cite{FsDetView} & - & - & - & - & 12.5 & 27.3 & 9.8 & 25.5 & 14.7 & 30.6 & 12.2 & 28.4\\	
 	FSOD-RPN~\cite{FSOD-RPN} & - & - & - & - & 12.0 & 22.4 & 11.8 & 30.8 & - & - & - & -\\
 	FSCE~\cite{FSCE} & - & - & - & - & 11.1 & - & 9.8 & - & 15.3 & - & 14.2 & - \\
 	DCNET\cite{DCNET} & - & - & - & - & 12.8 & 23.4 & 11.2& - & 18.6 & 32.6 & 17.5 & -\\
 	CME~\cite{CME} & - & - & - & - & 15.1 & 24.6 & 16.4 & - & 16.9 & 28.0 & 17.8 & - \\
 	DeFRCN~\cite{defrcn} & 16.1 & - & - & - & 18.5 & - & - & - & 22.6 & - & - & - \\
 	%\Xhline{0.5}
 	%Meta-DETR~\cite{MetaDetR} & 15.4 & 25.0 & 15.8 & - & 19.0 & 30.5 & 19.7 & - & 22.2 & 35.0 & 22.8 & - \\
	\bottomrule
	\end{tabular}
	}
	\end{center}
	\vspace{-2em}
\end{table*}

%\vspace{-1em}
\subsection{Gaussian Prototyping}
Furthermore, we propose a data augmentation scheme on the support features during meta-testing. Due to the limited number of shots per class in meta-testing, we notice that a naive averaging of the $K$-support shot features does not depict the true class prototype distribution, and hence the class prototypes are not as diverse as the ones during meta-training. Additionally, a large variance could exist between the $K$-shots, limiting the generalization of the class prototype. To address this issue, we assume that the support feature representation forms a class-conditional Gaussian distribution. To simulate this distribution, we compute the mean feature $\Bar{f}$ over the $K$-shots, and their standard deviation, $\sigma_f$. Then, we sample a latent vector $z$ from the Gaussian distribution $\mathcal{N}(\Bar{f}, \sigma_f^2)$, which becomes the class prototype, $\mu_c$. This augmentation strategy seeks to prevent overfitting on the novel support data. 

\begin{table*}[t!]
	\addtolength{\tabcolsep}{3pt}
	\begin{center}
	\caption{{ Evaluation of few-shot object detection on PASCAL VOC.} 
	The mean AP with IoU threshold 0.5 (AP50) on the 5 novel categories is reported for all the three different splits with $K=1,2,3,5,10$-shot. }
	\label{tab:DetVOC}
	\scalebox{0.75}{
	\begin{tabular}{l | ccccc | ccccc | ccccc}
	\toprule
	\multirow{2}{*}{Method}& \multicolumn{5}{c|}{Novel Set 1} & \multicolumn{5}{c|}{Novel Set 2} & \multicolumn{5}{c}{Novel Set 3} \\
	& 1 & 2 & 3 & 5 & 10 & 1 & 2 & 3 & 5 & 10 & 1 & 2 & 3 & 5 & 10 \\
	\midrule
	MetaYOLO~\cite{FSRW} & 14.8 & 15.5 & 26.7 & 33.9 & 47.2 & 15.7 & 15.2 & 22.7 & 30.1 & 40.5 & 21.3 & 25.6 & 28.4 & 42.8 & 45.9 \\
    MetaYOLO-CME~\cite{CME} & 17.8 & 26.1 & 31.5 & 44.8 & 47.5 & 12.7 & 17.4 & 27.1 & 33.7 & 40.0 & 15.7 & 27.4 & 30.7 & 44.9 & 48.8 \\
	MetaDet-YOLO~\cite{MetaDet} & 17.1 & 19.1 & 28.9 & 35.0 & 48.8 & 18.2 & 20.6 & 25.9 & 30.6& 41.5 & 20.1 & 22.3 & 27.9 & 41.9 & 42.9\\
	\rowcolor{HL}
	FSRN & \textbf{19.7} & \textbf{33.9} & \textbf{42.3} & \textbf{51.9} & \textbf{55.1} & \textbf{18.5} & \textbf{24.7} & \textbf{27.3} & \textbf{35.2} & \textbf{47.5} & \textbf{26.7} & \textbf{37.0} & \textbf{41.2} & \textbf{47.5} & \textbf{51.7} \\
	\midrule
	%LSTD~\cite{LSTD} & 8.2  & 1.0  & 12.4 & 29.1 & 38.5 & 11.4 & 3.8  & 5.0  & 15.7 & 31.0 & 12.6 & 8.5  & 15.0 & 27.3 & 36.3 \\
	%FSSP~\cite{MetaDet} & 41.6 & - & 49.1 & 54.2 & 56.5 & 30.5 & -& 39.5 & 41.4 & 45.1 & 45.1 & - & 36.7 & 45.3 & 49.4 & 51.3\\
    %Meta R-CNN~\cite{MetaRCNN} & 16.8 & 20.1 & 20.3 & 38.2 & 43.7 & 7.7 & 12.0 & 14.9 & 21.9 & 31.1 & 9.2 & 13.9 & 26.2 & 29.2 & 36.2 \\
    MetaDet~\cite{MetaDet}& 18.9 & 20.6 & 30.2 & 36.8 & 49.6 & 21.8 & 23.1 & 27.8 & 31.7 & 43.0 & 20.6 & 23.9 & 29.4 & 43.9 & 44.1\\
    FRCN-ft-full~\cite{TFA} & 15.2 & 20.3 & 29.0 & 25.5 & 28.7 & 13.4 & 20.6 & 28.6 & 32.4 & 38.8 & 19.6 & 20.8 & 28.7 & 42.2 & 42.1 \\
    TFA w/ fc~\cite{TFA}&  36.8 & 29.1 & 43.6 & 55.7 & 57.0 & 18.2 & 29.0 & 33.4 & 35.5 & 39.0 & 27.7 & 33.6 & 42.5 & 48.7 & 50.2 \\
    TFA w/ cos~\cite{TFA}&  39.8 & 36.1 & 44.7 & 55.7 & 56.0 & 23.5 & 26.9 & 34.1 & 35.1 & 39.1 & 30.8 & 34.8 & 42.8 & 49.5 & 49.8 \\
    MPSR~\cite{MPSR}&  42.8 & 43.6 & 48.4 & 55.3 & 61.2 & 29.8 & 28.1 & 41.6 & 43.2 & 47.0 & 35.9 & 40.0 & 43.7 & 48.9 & 51.3 \\
    FsDetView~\cite{FsDetView} & 25.4 & 20.4 & 37.4 & 36.1 & 42.3 & 22.9 & 21.7 & 22.6 & 25.6 & 29.2 & 32.4 & 19.0 & 29.8 & 33.2 & 39.8 \\ 
    FSCE~\cite{FSCE}&32.9 &44.0 &46.8 &52.9 &59.7 &23.7 &30.6 &38.4 &43.0 &48.5 &22.6 &33.4 &39.5 &47.3 &54.0\\
    CME~\cite{CME}& 41.5 &47.5 &50.4 &58.2 &60.9 &27.2 &30.2 &41.4 &42.5 &46.8 &34.3 &39.6 &45.1 &48.3 &51.5\\
    DCNET~\cite{DCNET}&33.9 & 37.4 &43.7 &51.1 &59.6 &23.2 &24.8 &30.6 &36.7 &46.6 &32.3& 34.9& 39.7& 42.6& 50.7\\
    DeFRCN\cite{defrcn}& 57.0 & 58.6 & 64.3 & 67.8 & 67.0 & 35.8 & 42.7 & 51.0 & 54.4 & 52.9 & 52.5 & 56.6 & 55.8 & 60.7 & 62.5 \\
	\bottomrule
	\end{tabular}}
	\end{center}
\vspace{-2em}
\end{table*}

\section{Experiments}
To evaluate our proposed model, we follow the well-established FSOD benchmarks \cite{FSRW, TFA, MPSR}, where experiments on MS-COCO \cite{coco}, and PASCAL VOC \cite{pascalvoc} datasets are conducted. We utilize the same classes and data splits as in previous works \cite{FSRW, TFA, MPSR} for fair comparisons. Due to space limitations, we provide the implementation details in the supplementary.
%\vspace{-0.5em}
\subsection{Datasets}

\textbf{MS-COCO.} The dataset comprises 80 classes, where 60 classes disjoint with VOC are used as base categories, while the remaining 20 are treated as novel classes. The $5k$ images from the validation set are utilized during meta-testing, and the rest for meta-training. We report the results for $K=5, 10, 30$-shots. 

\textbf{PASCAL VOC.} The dataset contains three different sets, where each one holds 20 classes. The classes are split randomly into 15 base and 5 novel classes. The data is sampled from both the VOC 2007 and VOC 2012 train/val sets for meta-training and meta-testing. For evaluation, the VOC 2007 test set is utilized for $K= 1,2,3,5,10,30$-shots.

\subsection{Results}
\textbf{Results on MS-COCO.} The results of our proposed approach are reported in Tab. \ref{tab:DetCoco}. We split the table to start with the one-stage FSOD methods~\cite{FSRW, MetaDet, once} followed by the two-stage based approaches~\cite{LSTD, MetaDet, MetaRCNN, TFA, MPSR, FsDetView, FSOD-RPN, FSCE, DCNET, CME, defrcn}. In comparison to meta-detectors, not only does FSRN outperform the dense meta-detectors by a significant margin, but also it delivers a superior performance than many sparse meta-detectors like \cite{MetaRCNN, FsDetView, FSOD-RPN} and is on par with \cite{CME, DCNET}.

\textbf{Results on PASCAL VOC.} The novel performance of the FSOD models on the PASCAL VOC dataset is presented in Tab.~\ref{tab:DetVOC}. In the upper table section, the results of one-stage based FSOD approaches~\cite{FSRW, MetaDet, once} are reported along with our proposed FSRN, while the remaining section shows the results for the two-stage based FSOD methods~\cite{LSTD, MetaDet, MetaRCNN, TFA, MPSR, FsDetView, defrcn} reporting their performance on the PASCAL VOC dataset. The proposed FSRN achieves a new standard as a dense meta-detector by a considerable margin across the various shot settings. Moreover, compared to sparse meta-detectors~\cite{MetaRCNN, MetaDet, FsDetView}, FSRN demonstrates quite a competitive performance. %except for~\cite{defrcn}\vspace{-3em}

\textbf{Model complexity.} The inference time per image on a GeForce 1080 GPU for novel classes on MS-COCO for the FSOD-RPN and FSRN are \SI{0.92}{\s} and  \SI{0.49}{\s}, respectively. Moreover, the number of FLOPS for FSOD-RPN and FSRN are  \SI{178.8}{G} and \SI{100.4}{G} FLOPS, respectively. The number of parameters of FSOD-RPN and FSRN are \SI{55.2}{M} and \SI{36.4}{M}, respectively. This showcases the speed and effectiveness of the proposed model.

\begin{table*}[t!]%[bp]
\caption{Ablation study on $10$-shot MS-COCO. Every row represents an incremental contribution. }
\label{table:novel_abl}
\begin{center}
\scalebox{0.9}{
\begin{tabular}{ll|cccc|cccc|cccc}
	\toprule
  &\multirow{2}{*}{Model Configuration} & \multicolumn{4}{c}{Base Performance} & \multicolumn{4}{c}{Novel performance} & \multicolumn{4}{c}{Transferability}\\
  &  & bAP & bAP50 & bAP75 & bAR & nAP & nAP50 & nAP75 & nAR & PT & PT50 & PT75 & RT \\
  \midrule
\textbf{A} &  Vanilla FSRN & 17.7 & 27.8 & 19.1 & 24.1 & 5.7 & 10.8 & 5.2 & 20.2 & 0.32 & 0.39 & 0.27 & \textbf{0.84} \\  
\textbf{B} &  + MWST & 30.6 & 45.8 & 33.4 & 52.6 & 12.4 & 21.2 & 12.5 & 30.7 & 0.40 & 0.46 & 0.37 & 0.58 \\  
\textbf{C} & + Early MSF & \textbf{32.5} & \textbf{48.6} & \textbf{35.0} & \textbf{54.0} & 15.1 & 25.3 & 15.2 & 32.1 & 0.46 & 0.52 & 0.43 & 0.59  \\  
%\textbf{D} & + Max Margin & 15.4 & 26.4 & 15.8 & 34.1  \\
\textbf{D} & + MSDA & \textbf{32.5} & \textbf{48.6} & \textbf{35.0} & \textbf{54.0} & 15.4 & 25.7 & 15.9 & 33.1 & 0.47 & 0.53 & 0.45 & 0.61   \\  
\rowcolor[HTML]{EFEFEF}
\textbf{E} & + Gaussian Prototyping & \textbf{32.5} & \textbf{48.6} & \textbf{35.0} & \textbf{54.0} & \textbf{15.8} & \textbf{26.4} & \textbf{15.9} &\textbf{ 36.0} & \textbf{0.49} & \textbf{0.54} & \textbf{0.45} & 0.67 \\  
\bottomrule
\end{tabular}}
\end{center}
\vspace{-2em}
\end{table*}
% \begin{table*}[]
% %\addtolength{\tabcolsep}{3pt}
% 	\begin{center}
% 	\caption{{\bf Ablation study on data augmentations.}
% 	We report the mean Averaged Precision and mean Averaged Recall on the 20 novel classes of MS-COCO in 10-shot setting.
% 	}
% 	\label{tab:ablation_da}
%     %\vspace{-2mm}
%     \scalebox{0.9}{
% 	\begin{tabular}{c | c | c | c | c | cccc}
% 	\toprule
% 	\multirow{2}{*}{MSF} & \multirow{2}{*}{MWST} & \multirow{2}{*}{$\mathcal{L}_{MM}$} & \multirow{2}{*}{MSDA} & \multirow{2}{*}{GP} & \multicolumn{4}{c}{Novel Performance} \\
% 	&  & & & & nAP & nAP50 & nAP75 & nAR \\
% 	\midrule
% 	\xmark & \xmark & \xmark & \xmark & \xmark & 5.7 & 10.8 & 5.2 & 20.2 \\
% 	\xmark & \xmark & \xmark & \cmark & \xmark & 8.0 & 14.5 & 7.7 & 34.1 \\
% 	\xmark & \xmark & \xmark & \cmark & \cmark & 9.7 & 17.2 & 9.7 & 34.0 \\
% 	\hline
% 	%\cmark & \xmark & \cmark & \cmark & \cmark & 13.0 & 22.3 & 13.1 & 28.4 \\
% 	%\hline
% 	\cmark & \cmark & \xmark & \xmark & \xmark & 15.1 & 25.3 & 15.2 & 32.1 \\
% 	\cmark & \cmark & \xmark & \cmark & \cmark & 15.4 & 25.7 & 15.9 & 33.1 \\
% 	\hline
% 	\cmark & \cmark & \cmark & \xmark & \xmark & 15.4 & 26.4 & 15.8 & 34.1 \\
% 	\cmark & \cmark & \cmark & \cmark & \cmark &{\bf 15.8} & {\bf 26.4} & {\bf 15.9} & {\bf 36.0}  \\
% 	\bottomrule
% 	\end{tabular}}
% 	\end{center}
%  \vspace{-2em}
% \end{table*}

\begin{table}[]
%\addtolength{\tabcolsep}{3pt}
	\begin{center}
	\caption{{\bf Ablation study on data augmentations.}
	We report the mean Averaged Precision and mean Averaged Recall on the 20 novel classes of MS-COCO in 10-shot setting.
	}
	\label{tab:ablation_da}
    %\vspace{-2mm}
    \scalebox{0.75}{
	\begin{tabular}{c | c | c | c | c | cccc}
	\toprule
	\multirow{2}{*}{MSF} & \multirow{2}{*}{MWST} & \multirow{2}{*}{$\mathcal{L}_{MM}$} & \multirow{2}{*}{MSDA} & \multirow{2}{*}{GP} & \multicolumn{4}{c}{Novel Performance} \\
	&  & & & & nAP & nAP50 & nAP75 & nAR \\
	\midrule
	\xmark & \xmark & \xmark & \xmark & \xmark & 5.7 & 10.8 & 5.2 & 20.2 \\
	\xmark & \xmark & \xmark & \cmark & \xmark & 8.0 & 14.5 & 7.7 & 34.1 \\
	\xmark & \xmark & \xmark & \cmark & \cmark & 9.7 & 17.2 & 9.7 & 34.0 \\
	\hline
	%\cmark & \xmark & \cmark & \cmark & \cmark & 13.0 & 22.3 & 13.1 & 28.4 \\
	%\hline
	\cmark & \cmark & \xmark & \xmark & \xmark & 15.1 & 25.3 & 15.2 & 32.1 \\
	\cmark & \cmark & \xmark & \cmark & \cmark & 15.4 & 25.7 & 15.9 & 33.1 \\
	\hline
	\cmark & \cmark & \cmark & \xmark & \xmark & 15.4 & 26.4 & 15.8 & 34.1 \\
	\cmark & \cmark & \cmark & \cmark & \cmark &{\bf 15.8} & {\bf 26.4} & {\bf 15.9} & {\bf 36.0}  \\
	\bottomrule
	\end{tabular}}
	\end{center}
 \vspace{-2em}
\end{table}
\subsection{Ablation Study}
%\vspace{-1em}
\textbf{Impact of proposed modules.} We conduct extensive experiments to study the effect of individual modules and their interactions. All experiments are performed on the MS-COCO dataset. In Table~\ref{table:novel_abl}, the performance on the base classes (bAP) is reported \textit{after} the meta-training phase, to showcase how the overall discriminability of the model is affected by the different components. We also report the performance on the novel classes (nAP) and the transferability. We present the results in an incremental way. In configuration A, we start with a direct extension of the meta-learning paradigm on RetinaNet. This version (vanilla FSRN) features a fusion mechanism directly before the detection head similar to Meta-Yolo and the RPN of FSOD-RPN. We find that this configuration has almost the same nAP as Meta-Yolo (5.6 in Table~\ref{tab:transferability}) but a higher bAP, which is attributed to the effect of the focal loss in RetinaNet. Adding the proposed MWST algorithm siginificantly boosts all metrics by almost doubling the bAP and nAP, and improving transferability. The proposed early fusion further boosts all metrics, especially the nAP. MSDA and Gaussian prototyping are only conducted in meta-testing and thus have no effect on the bAP. Their effect is reflected on the nAP and transferability. We present more ablation studies in the supplementary material.

\textbf{Effect of data augmentations.} In Table~\ref{tab:ablation_da}, a study on the impact of data augmentations, namely the multi-scale data augmentation (MSDA) and Gaussian Prototyping (GP), is conducted. Firstly, we show that without the multi-scale fusion (MSF), multi-way support tarining startegy (MWST), and max-margin loss ($\mathcal{L}_{MM}$) the effect of MSDA and GP are significant. When applying MSDA to the vanilla FSRN, we notice an nAP increase of $2.3$ points and a further boost of $1.3$ points when the GP is applied. Secondly, when MWSt, MSF, and/or $\mathcal{L}_{MM}$ is introduced without data augmentations, a significant jump in nAP is observed. This means that the aforementioned modules strengthen the discriminability of the proposed model. Hence, when applying the data augmentations we still witness an increase in nAP, yet marginal ($\sim 0.4$) points. The best performance is achieved when all the aforementioned modules contribute to the meta-testing phase as shown in the last row of Table~\ref{tab:ablation_da}. 

\begin{table}[t!]
	\addtolength{\tabcolsep}{5.0pt}
	\begin{center}
	\caption{{\bf Receptive field effect.}
	We report the mean Averaged Precision and mean Averaged Recall on the 20 novel classes of MS-COCO in 10-shot setting.}
	\label{tab:ablation_receptivefield}
    %\vspace{-2mm}
    \scalebox{0.9}{
	\begin{tabular}{c | cccc}
	\toprule
	Receptive Field & \multicolumn{4}{c}{Average Precision/Recall} \\
	/ Biggest Anchor & AP & AP50 & AP75 & AR \\
	\midrule
	3/10 & 12.4 & 21.2 & 12.5 & 30.7\\
	7/10 & 13.8 & 23.6 & 14.2 & 29.6\\
	11/10 & {\bf 15.8} & {\bf 26.4} & {\bf 15.9} & {\bf 36.0}  \\
	13/10 & 13.7 & 22.9 & 13.9 & 34.0  \\
	\bottomrule
	\end{tabular} }
	\end{center}
	\vspace{-2em}
\end{table}
\textbf{Effect of the Post-Fusion Receptive Field.} To measure the effect of the receptive field on the detection performance, we change the position of feature fusion in the network without changing the learning capacity. Specifically, we choose to fuse the features after different layers in the classification subnet. Fusing just before the classification head reduces the post-fusion receptive field to $3 \times 3$, while fusing before the entire subnet ($5$ conv-layers) results in a receptive field of $11 \times 11$. Table~\ref{tab:ablation_receptivefield} shows that optimal results are obtained when the post-fusion receptive field covers the biggest anchor size ($10 \times 10$). The AP drops as the receptive field decreases. We also experiment with $6$ layers after the fusion to test whether an increased model capacity improves the precision. However, it degrades the performance, highlighting the role of the post-fusion receptive field as a more important design parameter.

\textbf{Multiple runs.} All experiments are reported for seed 0 to have a fair comparison with other benchmarks. Due to time and resources limitations, we conducted multiple runs experiment on 10-shot MS-COCO benchmark, following TFA~\cite{TFA} and FSDetView~\cite{FsDetView}. Our model achieves an nAP of \textbf{$14.96 \pm 0.5$} which is better than the two previously mentioned models.

\subsection{Discussion and Limitations}
FSRN shows good generalization results on the challenging MS-COCO and PASCAL VOC. One limitation of our framework is that the MWST incurs extra computational cost during the training through the processing of more support images. Additionally, the training is sensitive to hyperparameters. Further work could address training stability of meta-detectors and the memory footprint of data augmentation techniques. We believe these contributions can be applied to other one-stage detectors or two-stage models. However, the extension to other detectors is out of the scope of this paper and is considered future work.

\section{Conclusion}
In this work, we have unveiled that the main underlying limitation of one-stage meta-detectors is their low discriminability and not their transferability. We found two factors adversely affecting discriminability. The first is that the post-fusion network provides a small receptive field that does not cover the anchor area and cannot effectively learn the class-specific instance-level features. As a remedy, a multi-scale fusion feature of the RetinaNet with an increased number of anchors was introduced. The second drawback stems from the scarcity of foreground samples per query image, inhibiting the learning signal. To alleviate this issue, we devised a multi-way support training strategy to leverage a higher number of annotations per query in a contrastive manner. A multi-scale data augmentation technique was proposed, accompanied by a reweighting of the foreground samples in the focal loss. We set a new standard for one-stage meta-detectors on MS-COCO and VOC datasets. 
%However, this extension is not trivial due to the extra incurred computational cost of the post-fusion detection subnet. 
The introduced improvements have shed more light on the meta-detection task and can be extended to two-stage meta-detectors for an enhanced region proposals generation.   

\clearpage

{\small
\bibliographystyle{ieee_fullname}
\bibliography{egbib}
}

\clearpage

\end{document}